# Arabic Offensive Language Detection Using Machine Learning and Ensemble Machine Learning Approaches

**Fatemah Husain**
Kuwait University, Department of Information Science, State of Kuwait
f.husain@ku.edu.kw

## Abstract

Ensemble machine learning is a meta-learning machine learning method that aims to improve single learner classifier's performance by combining predictions from multiple single learner classifiers. This study aims at investigating the effect of applying single learner machine learning approach and ensemble machine learning approach for offensive language detection on Arabic language. Classifying Arabic social media text is a very challenging task due to the ambiguity and informality of the written format of the text. Arabic language has multiple dialects with diverse vocabularies and structures, which increase the complexity of obtaining high classification performance. Our study shows significant impact for applying ensemble machine learning approach over the single learner machine learning approach. Among the trained ensemble machine learning classifiers, bagging performs the best in offensive language detection with F1 score of 88%, which exceeds the score obtained by the best single learner classifier by 6%. Our findings highlight the great opportunities of investing more efforts in promoting the ensemble machine learning approach solutions for offensive language detection models.

## 1. Introduction

The increasing number of online platforms for user generated content enables more people to experience freedom of expression than ever before. In addition, users of these platforms have the option of being anonymous and hiding their personal identity, which can increase the chance of misusing these technical features. Using offensive language has become one of the most common problems on social networking platforms. Text that contains some form of abusive behavior exhibiting actions with the intention of harming others is known as offensive language. Offensive language on social networking platforms can take multiple forms. Hate speech, aggressive content, cyberbullying, and toxic comments are different forms of offensive contents (Schmidt & Wiegand, 2017).

Reviewing the previous research on online offensive language detection, we find limited research covering Arabic language (Abdelfatah, Terejanu, & Alhelbawy (2017); Abozinadah, Mbaziira, & Jones (2015); Alakrot, Murray, & Nikolov, 2018; Albadi, Kurdi, & Mishra, 2018, 2019a; Haidar, Chamoun, & Serhrouchni, 2017, 2018, 2019; Johnston & Weiss, 2017 ; Kaati et al., 2015; Magdy, Darwish, & Weber, 2015; Mohaouchane, Mourhir, & Nikolov, 2019; Mubarak, Darwish, & Magdy, 2017). Furthermore, most of the literature has limitations and their scopes are not covering the topic of offensive language detection on a comprehensive basis. For example, Alakrot, Murray, and Nikolov (2018) focuses on offensive and abusive language in Arabic comments in YouTube, but their analysis depends on a very small dataset of 1,100 comments, which is not enough to generalize their findings. Another example, studies from Albadi, Kurdi, and Mishra (2018, 2019a) are very specific in scope for religious hate speech only in Arabic Twitter content. The majority of NLP researchers explore methods and techniques for automatic detection of offensive language in English text that cannot be generalized to Arabic and other languages that have different structures and rules from the English language.

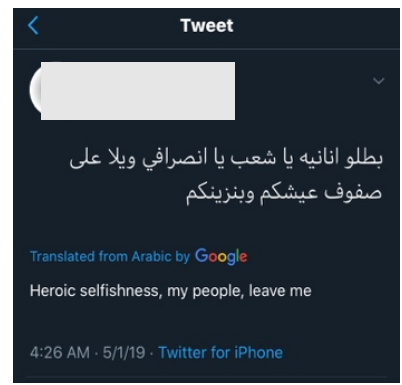

Figure 1: Example of an offensive language tweet.

Detecting offensive language for Arabic contents is a complex task. It has multiple challenges including: a) the informal language used in posts of social media which are usually written using short forms and slangs that are difficult to semantically process and understand by the classifier; b) The variation and diversity of the Arabic language dialects and forms that add difficulties to the task of identifying offensive contents as the texts might need to go through multiple preprocessing steps before feeding it into the classification model. To address the problem of informal language, we preprocess each tweet by converting emoticons and emojis to an Arabic textual description of its contents and segmenting hashtags into space separated words. The variation of Arabic dialects is addressed by converting dialectal Arabic to Modern Standard Arabic (MSA). The classifiers we experiment with include: machine learning models such as Support Vector Machine (SVM), logistic regression, and decision tree; ensemble machine learning models such as bagging, AdaBoosts, and random forest. We explore word-level features and character-level features.

In the rest of this paper, we organize the content as follows: section 2 discusses related work of Arabic offensive language detection on social media; section 3 introduces data description, details of preprocessing, and the methodology of our models; experimental results are discussed in section 4. We also present the conclusion of our work at the end of the paper.

## 2. Related Work

There are few studies that focus on detecting offensive Arabic tweets for identifying abusive Twitter accounts (Abozinadah, Mbaziira, and Jones, 2015; Abozinadah and ones, 2017; Abozinadah, 2017). Abozinadah, Mbaziira, and Jones (2015) construct an initial dataset starting from 500 Twitter accounts based on a set of Arabic swear words. Then, they check the most recent 50 tweets, profile pictures, and hashtags for each of these 500 Twitter accounts to reach a dataset of 350,000 Twitter accounts and 1,300,000 tweets with balanced classes, half labelled abusive and the other half labelled non- abusive. They use three types of features including profile-based features, tweet-based features, and social graph features to train three classifiers; Naïve Bayes (NB), SVM, and Decision Tree (J48). Results show that the NB outperforms the other classifiers when used with 100 features and 10 tweets for each account with an accuracy score of 85% (Abozinadah, Mbaziira, and Jones, 2015; Abozinadah, 2017).

Arabic language has been studied also by Alakrot, Murray, and Nikolov (2018a, 2018b) for automatic detection of offensive language. They construct a dataset from YouTube comments based on selecting channels that has controversial videos about celebrities. Their final dataset includes 167,549 comments posted by 84,354 users, and 87,388 replies posted by 24,039 users from 150 YouTube videos (Alakrot, Murray, and Nikolov, 2018a). Two classes of labels used: positive for offensive comments and negative for not offensive ones (Alakrot, Murray, and Nikolov, 2018a). They train an SVM classifier using two features; character n-gram (n= 1-5) and word-level feature. Results show best performance when using the SVM classifier with 10-fold cross validation and word-level features with 90.05% accuracy score (Alakrot, Murray, & Nikolov, 2018b).

Mohaouchane, Mourhir, and Nikolov (2019) explore multiple deep learning models to classify offensive Arabic language for YouTube comments using the same dataset developed by Alakrot, Murray, and Nikolov (2018a). They create word embedding using AraVec, which is trained on Twitter dataset and skip-gram model. Four deep learning models were evaluated for classifying offensive comments including convolutional neural network (CNN), Bidirectional Long short-term memory (Bi-LSTM), Bi-LSTM with attention mechanism, and combined CNN and LSTM. Results demonstrate an overall better performance for CNN with highest accuracy score of 87.84%, precision score of 86.10%, and F1 score of 84.05%, while the combined CNN-LSTM model shows better recall score of 83.46% (Mohaouchane, Mourhir, and Nikolov, 2019).

Ensemble machine learning methods have been applied to some applications of Arabic offensive language detection, such as cyberbullying. Haidar, Chamoun, and Serhrouchni (2017, 2019) use a dataset of 31,891 unbullying tweets and 2,999 bullying tweets manually labeled using two classes, 'bull' for bullying instances and 'None' for other instances. An algorithm for generating word embedding was used to create features. They investigate cyberbullying detection in two studies; one using single learner machine learning methods and another one using ensemble machine learning methods. The Naïve Bayes and the SVM classifiers were used in the machine learning single learner study. On the ensemble machine learning study, three models were explored: 1) stacking with simple linear regression as the meta-learning mechanism and classifiers include random forest, SVM, K-nearest neighbor, Bayesian logistic regression, and stochastic gradient descent; 2) three single learners with boosting as the meta-learning mechanism and classifiers include NB, SVM, and nearest neighbor; 3) similar to (2) but with bagging. The best result of the first study shows the NB single learner achieved F1 score of 90.05% (Haidar, Chamoun, and Serhrouchni, 2017), while the best results of the ensemble meta-learner shows 92.6% of F1 score (Haidar, Chamoun, and Serhrouchni, 2019). Thus, it is worth to try applying an enhancement to machine learning classifiers with ensemble meta-learning methods for this domain of problems.

In this study, our focus is on Natural Language Processing (NLP) techniques using machine learning approach and ensemble machine learning approach to analyze offensive language and hate speech in Arabic content of Twitter. Unlike previous Arabic offensive language detection studies, we do not incorporate Twitter user accounts into the classification model (Abozinadah, Mbaziira, and Jones, 2015; Abozinadah and ones, 2017; Abozinadah, 2017) and we do not implement deep learning models as earlier studies have done (Mohaouchane, Mourhir, and Nikolov, 2019). The data used in this study contains various offensive contents rather than limiting the content to specific source as Alakrot, Murray, and Nikolov (2018a, 2018b) which could narrow the types of offensive in samples into the context of the channel. For example, if the YouTube channel is a sport channel, then, the types of offensive language in the dataset reflect the nature of sport related offensive language. We train our models using linguistic features only. Moreover, the preprocessing steps we follow in this study are not identical to any of the preprocessing steps of the previous studies.

## 3. Data and Methodology

### 3.1 Data Description

We use the dataset provided by the shared task of the fourth workshop on Open-Source Arabic Corpora and Corpora Processing Tools (OSACT) in Language Resources and Evaluation Conference (LREC) 2020. The main goal of this shared task is to identify and categorize Arabic offensive language in Twitter. The organizers collect tweets through Twitter API and annotated them hierarchically regarding offensive language and offense type. The task is divided into two sub-tasks: a) detecting if a post is offensive or not offensive; b) identifying the offense type of an offensive post as hate speech or not hate speech. In addition, provider of the dataset performs some preprocessing to ensure the privacy of users. Twitter user mentions were substituted by

'@USER', URLs have been substitute by 'URL', and empty lines were replaced by '<LF>'.

The shared task releases the dataset into three different parts, training dataset, development dataset and testing dataset. We couldn't use the testing dataset in this study because the testing dataset version that is released on the time of this study is without labels. Thus, we split the training dataset into 20-80%, with 80% training tweets and 20% development tweets. The provided development dataset was used as the testing dataset. The summary of datasets distribution is presented in Table1. From the table, it is observable that the distributions of the datasets are twisted toward the not offensive class which makes the tasks much more complicated.

| Labels | Training | Development | Testing |
|---|---|---|---|
| Not Offensive | 5,191 | 1,295 | 821 |
| Offensive | 278 | 72 | 179 |
| Total Tweets | 5,468 | 1,367 | 1,000 |

Table 1: Datasets distribution.

## 3.2 Preprocessing

### 3.2.1 Emoji and Emoticon Conversion

To have a more robust system that is able to scale and cover more emojis and emoticon, we extract the entire set of emojis defined by Unicode.org (Unicode Organization, n.d.). Beautifulsoup4 4.8.2, a Python package for parsing HTML and XML documents, was used to scrape the emoji list available at the Unicode Organization website. We extracted the Unicode and name of each emoji using Beutifulsoup4. The emoji list contains 1,374 emojis. The Unicode Organization website provides textual descriptions for each emoji written in English. We extract these textual descriptions and then use translate 1.0.7 python package to translate them into Arabic language. Thus, during preprocessing of the data, each emoji is converted to its Arabic description. For example, 😋 is replaced by "وجه يستطعم الطعام". In addition, we manually develop emoticons list that includes a total of 140 emoticons with their textual descriptions in Arabic language. For example, ":-)" is replaced by "مبتسما". We analyze the description phrase as regular textual phrase in tweets thus it could maintain their semantic meanings after removing the original emoji and emoticon.

### 3.2.2 Arabic Dialects Normalization

The Arabic dialects has various forms differ, primarily, based on geography and social class (Habash, 2010). Arabic dialects are the Arabic languages that often been used in user-generated contents such as Twitter. Habash (2010) categorizes the Arabic dialects into seven dialects; Gulf, Egyptian, Iraqi, North African, Yemenite, Levantine, and Maltese Arabic that is not always considered one of the Arabic dialects. For example, the word "عافية - Afiah" means health in Gulf, Egyptian, Iraqi, and Levantine dialects, while it means fire in Moroccan Arabic. We reduce the dimensionality of the data by normalizing the variation of dialects on a set of nouns to be converted from dialectal Arabic to Modern Standard Arabic (MSA). For example, the variations of the word boy, 'رجل', 'زلمة', and 'زول' are converted to "ولد".

### 3.2.3 Letters Normalization

Arabic letters can be written in various format depending on the location of the letter within the word. We normalize Alif (إ،أ،ا to ا ), Alif Maqsura (ي،ئ to ى ), and Ta Marbouta (ة to ه). Letters that were repeated more than two times within a word were reduced to two times only.

### 3.2.4 Hashtag Segmentation

Hashtags are commonly used in Twitter to highlight important phrases within the tweet. Thus, it is very important to consider hashtags during the preprocessing phase to convert hashtags into a meaningful format. We remove the '#' symbol and replace '_' by a space. For example, the hashtag '#الهلال_التعاون' is converted to 'الهلال التعاون' which is easier for the system to understand and process.

### 3.2.5 Miscellaneous

Tweets were filtered to remove numbers, HTML tags, more than one space, and some symbols (e.g., " _", "'", "\"", "s", "...", "!", "?", "I", "@USER", "USER", "URL", ".", ",", ":", "/", "\\", ", ", "#", "@", "$", "&", ")", "(", "\"). We borrow the list of Arabic stopwords defined by Alrefaie (2016) which contains 750 stopwords. Furthermore, we remove diacritics which were used in tweets that contain text from the holy Qur'an or poetry.

## 3.3 Feature Extraction

We use count and TF-IDF features. For count features, we apply unigram features only, while for TF-IDF we apply n-gram of 1-2 words and 2-5 characters. All features were implemented using Python scikit-learn library.

## 3.4 Methodology

We use two approaches in our study. The first approach is based on machine learning models. Supervised machine learning models were used to classify the text as the actual labels are given. The second approach is based on ensemble machine learning models. Ensemble machine learning classifier is a meta-learning classifier, which combine predictions from multiple single classifiers; sometime called weak classifiers; with the objective of generating better prediction.

### 3.4.1 Machine Learning Models

We trained three machine learning classification models, namely Support Vector Machine (SVM), logistic regression and decision tree using the aforementioned features. We chose SVM because previous studies on offensive language detection show high performance when using an SVM based classifier (Abozinadah and Jones, 2017; Schmidt and Wiegand, 2017; Albadi, Kurdi, and Mishra, 2018). The logistic regression slightly differs from the SVM. The logistic regression concentrates on maximizing data probability, while the SVM focuses on maximizing the margin, the distance of the closest points to the hyperplane that separate between instances of classes using a linear function (Goodfellow, Bengio, and

Courville, 2016). Thus, we chose to examine the effect of this slight difference behind the algorithms of the SVM and the logistic regression by training a logistic regression classifier. We chose the decision tree to examine the classification task using a non-linear model, as the decision tree classifier uses a set of conditions to classify new instance (Eisenstein, 2018). We experimented with different parameter settings, but we only report the best performing settings. The Logistic regression classifier was trained using word and character n-gram features (n = 1-4) with L2 regularization. The SVM classifier was also trained using word and character n-gram features (n = 1-4) with linear kernel andL2 regularization. The decision tree classifier was trained using the same features of the previous classifiers with gini criterion and best splitter. We used Python scikit-learn library to implement all models.

### 3.4.2 Ensemble Machine Learning Models

We trained three ensemble machine learning classification models, namely bagging, random forest, and AdaBoost. We select three different models each use different ensemble machine learning method. The bagging model runs a learning algorithm multiple times, in which randomly selected samples from the dataset are given to the learning algorithm in each run (Nadali et al., 2013). The random forest uses a final decision of the average prediction that is given by a single decision tree within a combination of multiple decision trees (Haidar, Chamoun, and Serhrouchni, 2019). The AdaBoost is a boosting ensemble machine learning model that depends on a sequential learning of classifiers; the single classifiers are tweaked based on the misclassified samples from the previous classifiers (Kaati et al., 2015). The final decision of AdaBoost is the weighted sum of outputs from a combination of the final classifications (Kaati et al., 2015). The bagging classifier was trained with learning rate of 1 and 50 maximum number of estimators. The random forest was trained using 100 maximum number of trees and Gini criterion. The AdaBoost was trained using a learning rate of 1 and 50 maximum number of estimators. All classifiers were trained using word and character n-gram features. We used Python scikit-learn library to implement all models.

### 4. Experiment Results

Figure 2 and Figure 3 shows the results for performance evaluation using precision, recall, accuracy, and F1 metrics. One observation from the figures shows the problem of imbalanced data, so that higher accuracy does not guarantee higher F1 score. Thus, F1 score is more informative for evaluating the classifiers. Among the machine learning models, the SVM based classifier performs the best with F1 score of 82% followed by the logistic regression based classifier with F1 score of 81%, and lastly the decision tree classifier with F1 score of 69%. Among the ensemble machine learning models, the bagging based classifier performs the best with F1 score of 88% followed by the random forest based classifier with F1 score of 87%, and at the end comes the Adaboost based classifier with F1 score of 86%.

These results demonstrate the effectiveness of using ensemble machine learning methods; F1 score increases from 82% using the best machine learning model to 88% using the best ensemble machine learning model.

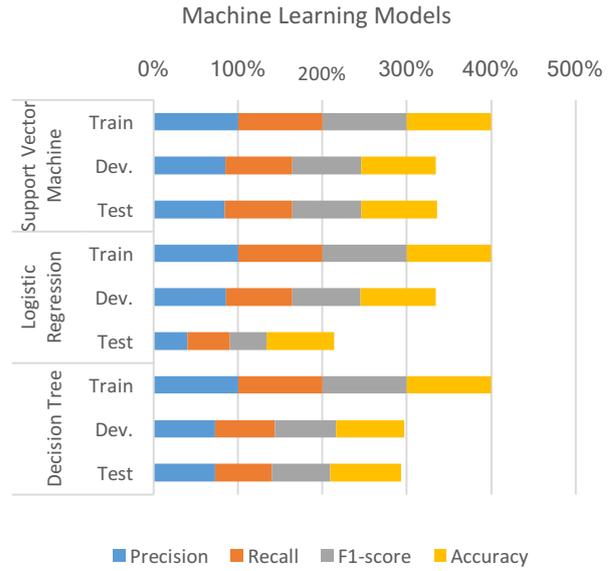

Figure 2: Performance evaluation results for machine learning models.

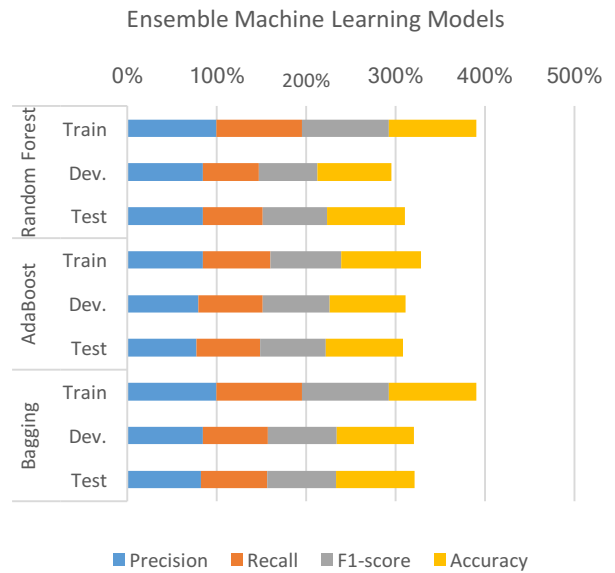

Figure 3: Performance evaluation results for ensemble machine learning models.

### 5. Conclusion

Ensemble machine learning is a meta-learning machine learning method that aims to improve single learner classifier's performance by combining predictions from multiple single learner classifiers. In this study, we investigate the effect of applying single learner machine learning approach (SVM, logistic regression, and decision tree) and ensemble machine learning approach (bagging, Adaboost, and random forest) on offensive language detection for Arabic language. Online offensive language classification task is very challenging due to the ambiguity and informality of the social media language, which increases the difficulties to achieve high performance, particularly for Arabic text that has multiple dialects. Results show promising impact for the ensemble machine learning approach over the single learner machine learning approach. Among the trained ensemble machine learning classifiers, bagging performs the best in offensive language detection with F1 score of 88%, which exceeds the score obtained by the best single learner classifier by 6%.

## 6. Bibliographical References

Abdelfatah, K., Terejanu, G., and and Alhelbawy, A. (2017). Unsupervised Detection of Violent Content In Arabic Social Media. *Comput. Sci. Inf. Technol.* (CS IT), pp. 1–7, 2017

Abozinadah, E. , Mbaziira, A., and Jones, J. (2015). Detection of Abusive Accounts with Arabic Tweets. *International Journal of Knowledge Engineering*, Vol. 1, No. 2. DOI: 10.7763/IJKE.2015.V1.19

Abozinadah, E. (2017). Detecting Abusive Arabic Language Twitter Accounts Using a Multidimensional Analysis Model.

Abozinadah, E., and Jones, J. (2017). A Statistical Learning Approach to Detect Abusive Twitter Accounts. *Proceedings of the International Conference on Compute and Data Analysis*, 130280, 6–13. https://doi.org/10.1145/3093241.3093281

Alakrot, A., Murray, L., and Nikolov, N. S. (2018a). Dataset Construction for the Detection of Anti-Social Behaviour in Online Communication in Arabic. *Procedia Computer Science*, *142*, 174–181. https://doi.org/https://doi.org/10.1016/j.procs.2018.10.473

Alakrot, A., Murray, L., and Nikolov, N. S. (2018b). Towards Accurate Detection of Offensive Language in Online Communication in Arabic. *Procedia Computer Science*, *142*, 315–320. https://doi.org/https://doi.org/10.1016/j.procs.2018.10.491

Albadi, N., Kurdi, M., and Mishra, S. (2018). Are they Our Brothers? Analysis and Detection of Religious Hate Speech in the Arabic Twittersphere. 2018 IEEE/ACM International Conference on Advances in Social Networks Analysis and Mining (ASONAM), 69–76. https://doi.org/10.1109/ASONAM.2018.8508247

Albadi, N., Kurdi, M., and Mishra, S. (2019b). Hateful People or Hateful Bots?: Detection and Characterization of Bots Spreading Religious Hatred in Arabic Social Media. *Proceedings of the ACM on Human-Computer Interaction*, *3*(CSCW), 1–25. https://doi.org/10.1145/3359163

Alrefaie, M. (2016). Arabic-stop-words [Github Repository]. Retrieved on January 24, 2020 from https://github.com/mohataher/arabic-stop-words

Alshehri, A., Nagoudi, E.M., Alhuzali, H., and Abdul-Mageed, M. (2018). Think Before Your Click: Data and Models for Adult Content in Arabic Twitter.

Davidson, T., Warmsley, D., Macy, M., and Weber, I. (2017). Automated Hate Speech Detection and the Problem of Offensive Language. AAAI Publications, Eleventh International AAAI Conference on Web and Social Media About offensive content. arXiv.org. Retrieved from http://search.proquest.com/docview/2074118430/

Eisenstein, J. (2018) Natural Language Processing. MIT press. Retrieved from https://github.com/jacobeisenstein/gt-nlp-class/blob/master/notes/eisenstein-nlp-notes.pdf

Habash, N. (2010). Introduction to Arabic natural language processing (Vol. 3, pp. 1–185).

Haidar, B., Chamoun, M., and Serhrouchni, A. (2017). Multilingual cyberbullying detection system: Detecting cyberbullying in Arabic content. 2017 1st Cyber Security in Networking Conference (CSNet), 2017-, 1–8. https://doi.org/10.1109/CSNET.2017.8242005

Haidar, B., Chamoun, M., and Serhrouchni, A. (2018). Arabic Cyberbullying Detection: Using Deep Learning. 2018 7th International Conference on Computer and Communication Engineering (ICCCE), 284–289. https://doi.org/10.1109/ICCCE.2018.8539303

Haidar, B., Chamoun, M., and Serhrouchni, A. (2019). Arabic Cyberbullying Detection: Enhancing Performance by Using Ensemble Machine Learning. 2019 International Conference on Internet of Things (iThings) and IEEE Green Computing and Communications (GreenCom) and IEEE Cyber, Physical and Social Computing (CPSCom) and IEEE Smart Data (SmartData), 323–327. https://doi.org/10.1109/iThings/GreenCom/CPSCom/SmartData.2019.00074

Johnston, A., and Weiss, G. (2017). Identifying sunni extremist propaganda with deep learning. 2017 IEEE Symposium Series on Computational Intelligence (SSCI), 2018-, 1–6. https://doi.org/10.1109/SSCI.2017.8280944

Kaati, L., Omer, E., Prucha, N., and Shrestha, A. (2015). Detecting Multipliers of Jihadism on Twitter. 2015 IEEE International Conference on Data Mining Workshop (ICDMW), 954–960. https://doi.org/10.1109/ICDMW.2015.9

Mohaouchane, H., Mourhir, A., and Nikolov, N. S. (2019). Detecting Offensive Language on Arabic Social Media using Deep Learning, (December). https://doi.org/10.1109/SNAMS.2019.8931839

Mubarak, H., Darwish, K., and Magdy, W. (2017). Abusive Language Detection on Arabic Social Media. *Proceedings of the First Workshop on Abusive Language Online. Association for Computational Linguistics (ACL)*, 2017. pp. 52–56.

Nadali, S., Murad, M., Sharef, N., Mustapha, A. and Shojaee, S. (2013). A review of cyberbullying detection: An overview. *13th International Conference on Intellient Systems Design and Applications*, Bangi, pp. 325-330.